# Prediction of COPD Using Machine Learning, Clinical Summary Notes, and Vital Signs

Negar Orangi-Fard

Department of Mathematics and Statistics,
School of Science and Technology,
Georgia Gwinnett College,
Lawrenceville, GA 30043, USA
Email: norangifard@ggc.edu

*Abstract*— Chronic obstructive pulmonary disease (COPD) is a chronic inflammatory lung disease that causes obstructed airflow from the lungs. In the United States, more than 15.7 million Americans have been diagnosed with COPD, with 96% of individuals living with at least one other chronic health condition. It is the 4th leading cause of death in the country. Over 2.2 million patients are admitted to hospitals annually due to COPD exacerbations. Monitoring and predicting patient's exacerbations on-time could save their life. This paper presents two different predictive models to predict COPD exacerbation using AI and natural language processing (NLP) approaches. These models use respiration summary notes, symptoms, and vital signs. To train and test these models, data records containing physiologic signals and vital signs time series were used. These records were captured from patient monitors and comprehensive clinical data obtained from hospital medical information systems for tens of thousands of Intensive Care Unit (ICU) patients. We achieved an area under the Receiver operating characteristic (ROC) curve of 0.82 in detection and prediction of COPD exacerbation.

*Clinical Relevance*—In above two-thirds of patients at risk, COPD is underdiagnosed or misdiagnosed. The identification of patients with early-stage COPD would improve outcomes and reduce the burden on healthcare systems. Prediction models can identify patients with COPD who might benefit from earlier or more specific intervention. Our work suggests that clinically available patient's data and vital signs can predict risk of COPD exacerbations.

## 1.  INTRODUCTION

Chronic obstructive pulmonary disease (COPD) is a common disease characterized by persistent respiratory symptoms and airflow limitation. By 2020, COPD become a serious health concern, third rank in terms of mortality and fifth in terms of global burden. COPD causes more than 500,000 hospitalizations and more than 100,000 deaths in the United States every year. Chronic obstructive pulmonary disease is defined as FEV1/FVC, where FEV1 is forced expiratory volume in one second and FVC forced vital capacity. This ratio reflects the amount of air you can forcefully exhale from your lungs and is measured using spirometry [1,2,3].

A COPD exacerbation, or flare-up, occurs when the COPD respiratory symptoms become much more severe. A considerable percentage of COPD patients experience periodic exacerbations of symptoms which are serious threats to the patients. Exacerbation accelerate decline in lung functions resulting in permanent functional decline in the patients and are associated with increase in the rehospitalization risk and death [4].

In clinical practice, the predictor of frequent exacerbations defined as two or more exacerbations and one severe exacerbation per year is used to guide therapeutic choices for exacerbation prevention, but this approach is clinically limited. Predictive machine learning models created based on clinical databases are good approaches for making predictions and personalized decisions about the patients [5,6].

Matheson et al. [7] reviewed and assessed the performance of all published models that predicted development of COPD. The identified 4,481 records and selected 30 articles for full-text review none of the models had good accurately for predicting future risk of COPD.

Guerra et al. [8] studied and identified models that predict exacerbations in COPD patients. Out of 1382 studies, they included 25 studies with 27 prediction models and studied their performance in predicting exacerbation in patients with COPD.

None of the existing models fellfield the requirements to be used for predicting exacerbations in COPD patients. To address this and overcome this shortcoming, in this study a novel framework termed CPML with predictive models for COPD exacerbation prediction is proposed. It employs machine learning methods based on caregiver notes from clinical data and vital signs data which leads to better management of COPD patients.

## 2. METHODS

In this section, the proposed methodology for predicting COPD using unsupervised text mining and machine learning techniques is described.

### A. Data and Pre-Processing

The following datasets were used in this work:

(1) MIMIC-III Clinical Database [9]: data from the Medical Information Mart for Intensive Care database, which is a large, freely available database comprising de-identified health-related data associated with over forty thousand patients who stayed in critical care units at Beth Israel Deaconess Medical Center in Boston, Massachusetts between 2001 and 2012. MIMIC-III clinical database is a relational database consisting of 26 tables. Tables include information such as demographics, vital signs taken at the bedside, laboratory test results, procedures, medications, caregiver notes, imaging reports, and mortality. Tables are linked by hadm-id, which is a unique identifier for each admission. In this work ADMISSIONS and NOTEEVENTS tables from MIMICIII Clinical Database were used. The ADMISSIONS table contains admission information and has 58,976 rows and 19 columns. The columns for the NOTEEVENTS table have 2,083,180 rows which contain all notes for each hospitalization which links with hadm-id.

(2) MIMIC-III Waveform Database Matched Subset [10]: it contains 22,317 waveform records, and 22,247 numeric records, for 10,282 distinct ICU patients. These recordings typically include digitized signals such as ECG, ABP, respiration, and PPG, as well as periodic measurements such as heart rate, oxygen saturation, and systolic, mean, and diastolic blood pressure.

*B. CPML Framework and Predictive Models:*

Two predative models under a framework termed CPML were developed. **Figure 1** shows the block diagram for the CPML framework to predict flare-up in COPD patients. The data processing steps and models are explained in below:

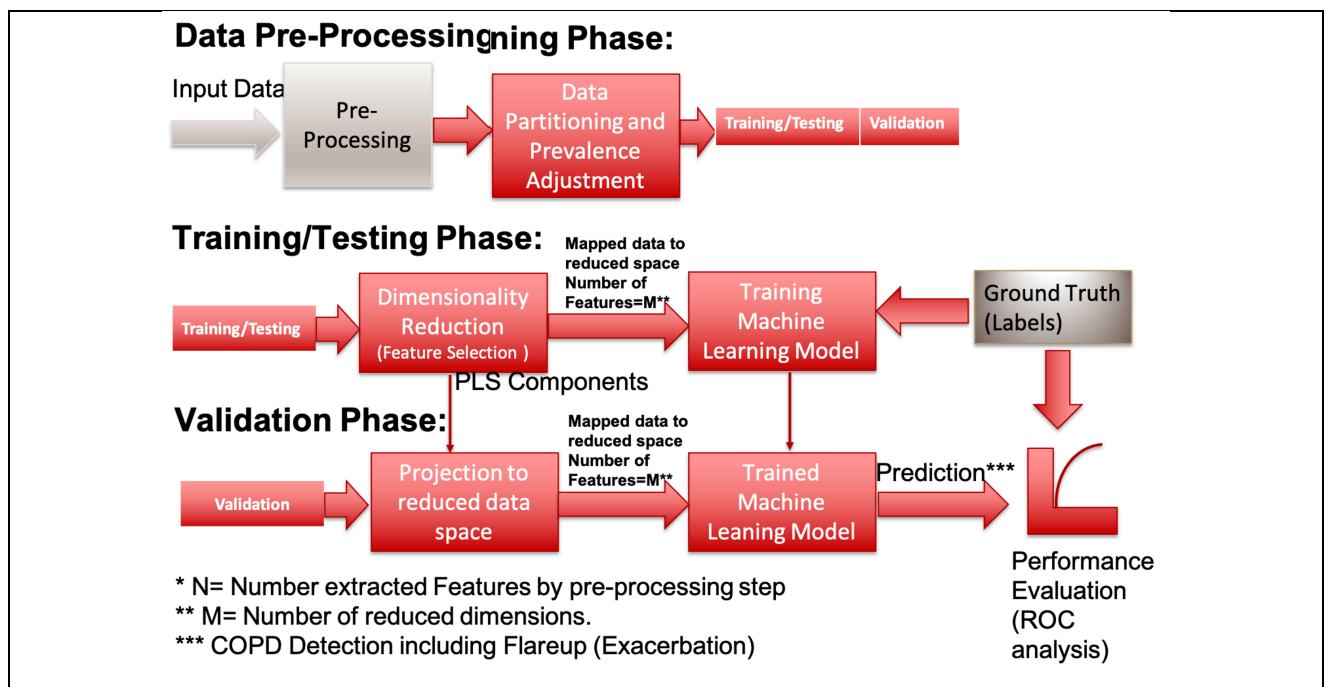

**Figure 1: CPML framework and its models to predict COPD.**

***B.1. Pre-Processing:*** This step based on model type is different as shown in **Figure 2**. In this work two following models were developed:

- *Model 1*:

COPD predictive model based on respiratory clinical notes: This model uses natural language processing (NLP) [11] to convert the clinical notes to numerical data and then feed them to a machine learning model for predicting COPD. A bag-of-words approach from natural language processing was used on the respiratory clinical notes. First, the notes were modified by removing newlines and carriage returns and replacing the missing text with space. Next, a tokenizer was built to split the note into individual words. A vectorizer on the clinical notes was built, where every row represents a different document, and every column represents a different word. The output is a space matrix called the "Document-Term-Matrix". A histogram of all words in this matrix was plotted to check the most frequently used words and identify words that do not add any value to the prediction model (words such as "the", "or", "and"), and they were added to the stop words list. A tokenizer with these new stop words were rebuilt and used to transform text notes in the dataset into the numerical data (features).

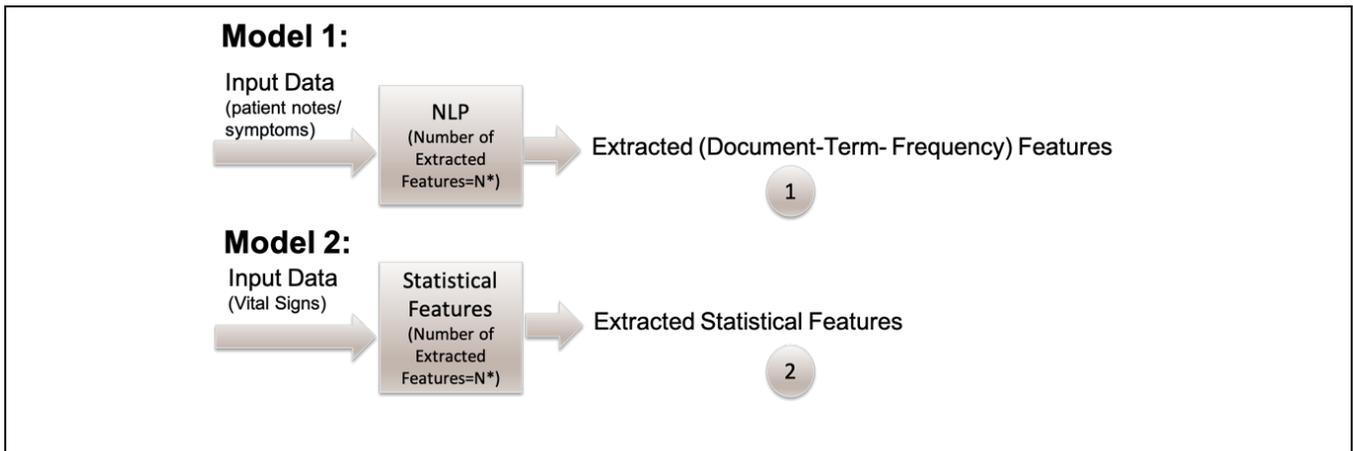

**Figure 2:** Pre-Processing step for each model in Figure 1.

- *Model 2:*

*COPD predictive model based on vital signs:* it uses vital signs (Heart Rate, SpO2 and Respiration Rate signals) and statistical features described in this section to extract meaningful features. For each of the Heart Rate, SpO2 and Respiration Rate signals the following statistical features were calculated: "Maximum", "Minimum", "Mean", "Median" and "Standard Deviation". Additional features were also extracted by bucketing the signals based on threshold values defined in Global Initiative for Chronic Obstructive Lung Disease (GOLD) staging definition [12]. The stages for the heart rate are

"Normal" for less than 90 beats per minutes (bpm), "Mild" for 90-100 bpm, "Moderate" for 100-110 bpm, "Severe" for 110-120 bpm and "Very Severe" for greater than 120 bpm. For Respiration Rate, the stages are "Normal" for 12-18 breath per minute, "Low" for less than 12 bpm, "High" for 18-20, "Abnormal" for greater than 20 breaths per minute. For SpO2, the stages are "Normal" for greater than 92%, "Mild" for 90%-92%, "Moderate" for 85%-90%, "Severe" for 80%-85% and "Very Severe" for less than 80%.

*B.2 Data Partitioning and Prevalence Adjustment:* To train and test the proposed machine learning model, the dataset should be partitioned to training and testing dataset. In case the number of none-COPD data records in the training set is much higher than COPD data records, data records were randomly selected out of none-COPD records to make sure there is a balance between COPD and none-COPD data records in the training set. This is called data balancing or prevalence adjustment. The leftover none-COPD data record in training set after prevalence adjustment were added to validation set.

*B.3 Dimensionality Reduction:* Passing all extracted features to machine learning (ML) dramatically increases computational time and it might impact the performance of machine learning model. When number of features to ML is very high, typically feature selection or dimensionality reduction is applied to obtain the most important features to pass to Machine Learning. In this work partial least-squares (PLS) regression [13] was used to eliminate redundant information and only pass the most important information to ML.

*B.4 Machine Learning Methods:* three different machine learning techniques as described below were used to predict COPD and their performance were compared:

- *Support vector machine (SVM)*:

SVM project the input features into a higher-dimensional space, where they can find an optimal hyperplane that maximizes the margin between the hyperplane and the closest data points. In this study, we used a Gaussian radial basis function (RBF) kernel to extend the SVM for non-linear classification [14].

- *Quadratic discriminant analysis (QDA)*:

QDA uses a transformation function to maximize the ratio of between-class variance to within-class variance and to minimize the overlap of the transformed distributions. A "pseudo-quadratic" (SQ) transformation was used. SQ uses an inverse covariance matrix as a cost function (how well the machine learning algorithm maps training data to outcomes) to measure the variability of covariance matrices among the classes [15].

- *Adaptive Boosting (AdaBoost):*

AdaBoost is an ensemble technique that combines multiple weak classifiers, or decision trees, that work in conjunction to reach the final classification decision. Linear discriminant analysis seeks to find a linear combination of features to separate classes. Although the decision boundary for linear discriminant analysis is a line, the boundary for quadratic discriminant analysis is a quadratic equation, which increases model flexibility with the tradeoff of greater complexity [16].

*B.5 Performance Evaluation and Validation:*

To evaluate the CPML framework performance, Receiver Operating Characteristic (ROC) analysis was done. The ROC curve represents the performance of a binary classifier as its discrimination threshold is varied [17]. It is a plot of the true positive rate (TPR) against the false positive rate (FPR) at various threshold settings. The true-positive rate is sensitivity, recall or probability of diagnosis. The false-positive rate is probability of false diagnosis and can be calculated 1 – specificity [17]. In this work, multiple receiver-operating-characteristic (ROC) curves were created for each predicative model and machine learning methods, and the area under ROC curves were estimated. The classification accuracies were also calculated to evaluate performance of CPML models.

## 3. RESULTS

In this section the performances of CPML framework and its predictive models were compared using data explained in the method section. The data were divided into training and validation sets using data partitioning and adjustment explained in the method section.

- Model 1 (COPD predictive model based on clinical notes) performance:

Input to this model were the respiratory clinical notes. 31667 records from "MIMIC-III Clinical Database" [9] were used, where 354 records were diagnosed as COPD patients. 50% (15833) of the data were used for the training and 50% of the data were used to test the CPML models. The clinical note data were passed to the NLP module to convert text data and extract meaningful features from the notes. A bag-of-words approach was used to process the discharge summary notes data. This resulted a Document-Term-Matrix with 3000 features. **Figure 3** shows the "word cloud" plot for all clinical notes used for training and testing the model 1. The extracted 3000 features were passed to the PLS module to eliminate redundant information and keep only the important features (n=15). These reduced features were passed to the machine learning models to predict COPD. The optimal performance was achieved using 15 PLS components. **Figure 4** shows corresponding ROC curves for model 1 and three different machine learning models to predict COPD using respiratory clinical notes and NLP. Table 1 shows the areas under ROC curve for those machine learning models on the validation set only.

**Figure 3: A word cloud of respiratory clinical notes for COPD patients.**

**Figure 4: ROC curves for model 1 using respiratory clinical note data, NLP and machine learning to predict COPD. The results for three different machine learning techniques (SVM, QDA, AdaBoost) are shown.**

TABLE 1

| Machine Learning Type | Accuracy | AUC |
|---|---|---|
| SVM | 84.0% | 0.82 |
| Ada-Boost | 78.2% | 0.79 |
| QDA | 75.0% | 0.77 |

- Model 2 (COPD predictive model based on vital signs) performance:

Input to this model were the vital signs (respiration rate, heart rate and SpO2). 10489 records for MIMIC-III Waveform Database Matched Subset [10] were used, where 2551 records were diagnosed as COPD patients. 70% (5591) of the data were used for the training and 30% of the data were used to test the CPML models. The vital signs were passed to the statistical features extraction module to extract meaningful features from the notes. The extracted were passed to the PLS module to avoid redundant information and keep only the important feature. The reduced features (n=15) were passed to the machine learning models to predict COPD. The optimal performance was achieved using 15 PLS components and 70%:30% data partitioning between training and validation set. **Figure 5** shows corresponding ROC curves for three different machine learning models to predict COPD using vital signs. **Table 2** shows the average area under ROC curve for those machine learning models on the validation set.

TABLE 2

| Machine Learning Type | Accuracy | AUC |
|---|---|---|
| SVM | 77.0% | 0.79 |
| Ada-Boost | 83.0% | 0.77 |
| QDA | 67.0% | 0.77 |

By comparing models 1 and 2 results, the model 1 using NLP and respiratory clinical notes outperformed compared to the model 2 using only vital signs data.

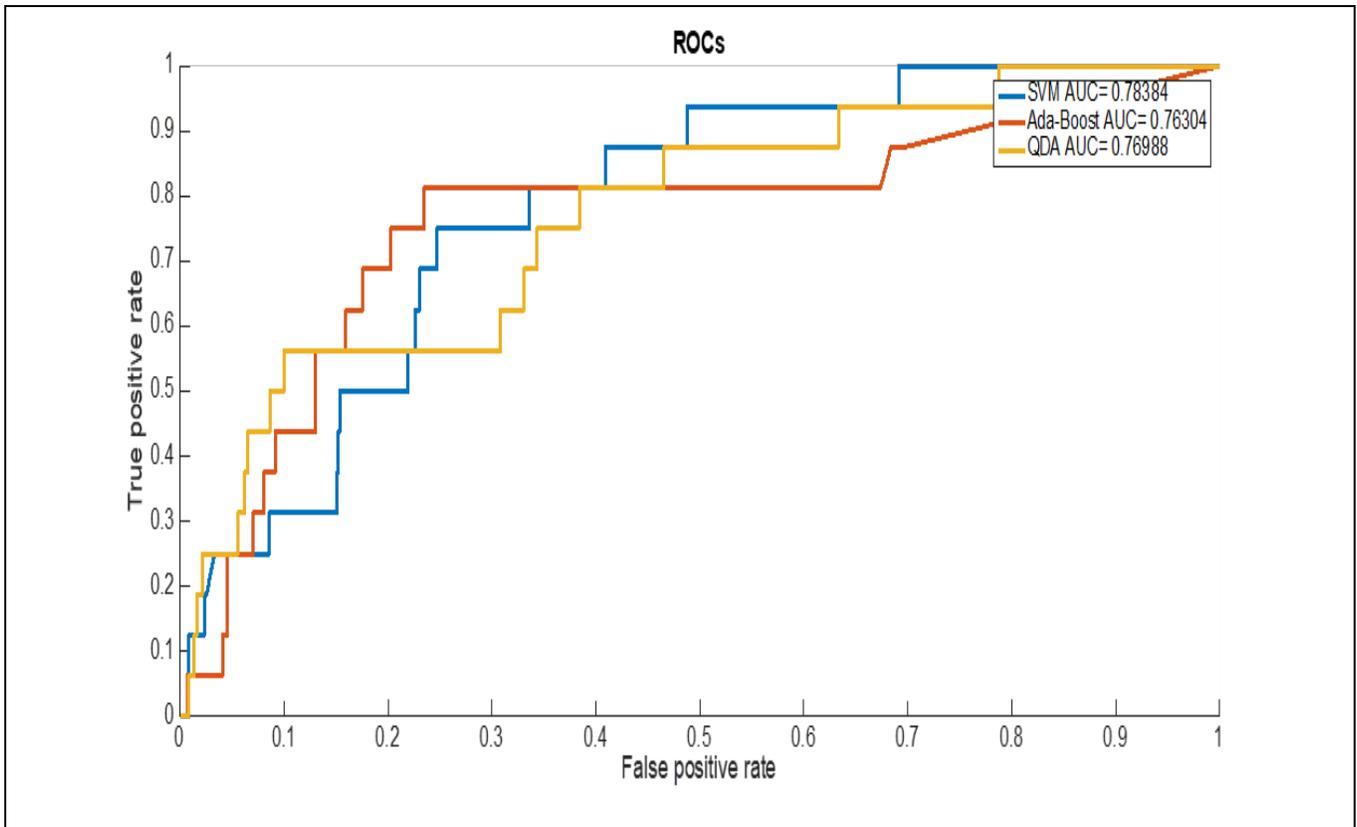

**Figure 5: ROC curves for model 2 using vital signs and machine learning to predict COPD. The results for three different machine learning techniques (SVM, QDA, AdaBoost) are shown.**

4. CONCLUSION

The information contained within respiratory clinical notes and vital signs, correspondingly in the MIMIC-III dataset [9] and MIMIC-III Waveform Database Matched Subset [10] were used to predict COPD, comparing the performance of three machine learning models, SVM, AdaBoost, and QDA. After selecting the most dominant and important features using PLS based dimensionality reduction, it was found that the SVM had the best performance with an AUROC curve of 0.82, followed by AdaBoost, and QDA.

This work was only covered using respiratory clinical notes and vital signs. However, there are many other types of data that could be used as well, including laboratory values, imaging reports, progress notes, ventilator settings, etc. It is reasonable to expect that a future work incorporating and fusing all this information could lead to models with even better performance. As part of future directions, additional data could certainly improve the accuracy of the model and could be also used to validate CPML on independent clinical datasets. Also, other machine learning approaches, such as deep learning could be employed. In addition, CPML framework could also be useful for other applications, such as predicting length of stay, readmission or patient survival.